\documentclass[10pt,twocolumn,letterpaper]{article}

\usepackage{icb}
\usepackage{times}
\usepackage{epsfig}
\usepackage{graphicx}
\usepackage{amsmath}
\usepackage{amssymb}
\usepackage{tabulary}
\usepackage{multirow}
\usepackage{booktabs}  
\usepackage{caption}
\usepackage{subfigure}
\usepackage{float}
\usepackage{fancyhdr}

\usepackage[pagebackref=true,breaklinks=true,letterpaper=true,colorlinks,bookmarks=false]{hyperref}

\icbfinalcopy 

\ificbfinal\fi
\begin{document}

\title{Improving Face Anti-Spoofing by 3D Virtual Synthesis}
\author{Jianzhu Guo\thanks{The authors make equal contributions.} $^{\,1,2}$, Xiangyu Zhu\footnotemark[1] $^{\,1}$, Jinchuan Xiao$^{1,2}$, Zhen Lei\thanks{Corresponding author.} $^{\,1}$, Genxun Wan$^{3}$, Stan Z. Li$^{1}$\\
\normalsize{$^1$NLPR, Institute of Automation, Chinese Academy of Sciences, Beijing, China} \\
\normalsize{$^2$University of Chinese Academy of Sciences, Beijing, China}\\
\normalsize{$^3$First Research Institute of the Ministry of Public Security of PRC}\\
{\tt\small \{jianzhu.guo, xiangyu.zhu, jinchuan.xiao, zlei, szli\}@nlpr.ia.ac.cn, friwan@163.com}
}

\maketitle
\thispagestyle{empty}

\begin{abstract}
   Face anti-spoofing is crucial for the security of face recognition systems. Learning based methods especially deep learning based methods need large-scale training samples to reduce overfitting. However, acquiring spoof data is very expensive since the live faces should be re-printed and re-captured in many views.
   In this paper, we present a method to synthesize virtual spoof data in 3D space to alleviate this problem.
Specifically, we consider a printed photo as a flat surface and mesh it into a 3D object, which is then randomly bent and rotated in 3D space. Afterward, the transformed 3D photo is rendered through perspective projection as a virtual sample.
The synthetic virtual samples can significantly boost the anti-spoofing performance when combined with a proposed data balancing strategy.
   Our promising results open up new possibilities for advancing face anti-spoofing using cheap and large-scale synthetic data.
\end{abstract}

\vspace{-1.5em}

\makeatletter 
\def\ps@IEEEtitlepagestyle{ 
\def\@oddfoot{\mycopyrightnotice} 
\def\@evenfoot{} 
} 
\def\mycopyrightnotice{ 
{\hfill \footnotesize 978-1-7281-3640-0/19/\$31.00 \copyright 2019 IEEE\hfill} 
} 
\makeatother 

\section{Introduction}

Due to the intrinsic distinctiveness and convenience of biometrics, biometric-based systems are widely used in our daily life for person authentication. The most common applications cover phone unlock (e.g., iPhone X), access control, surveillance, and security.
Face, as one of the biometric modalities, gains increasing popularity in academic and industry community~\cite{guo2017multi,guo2018dominant}. Face recognition has achieved great success in terms of verification and identification~\cite{kemelmacher2016megaface,guo2020learning,cao2020domain}.
However, spoof faces can be easily obtained by printers (i.e., print attack) and digital camera devices (i.e., replay attack). These spoofs can be very similar to genuine faces in appearance with proper handlings, like bending and rotating.
Therefore, it is important to equip the face recognition system with robust presentation attack detection (PAD) algorithms.

\begin{figure}[!htb]
  \small
  \centering
  \includegraphics[width=0.45\textwidth]{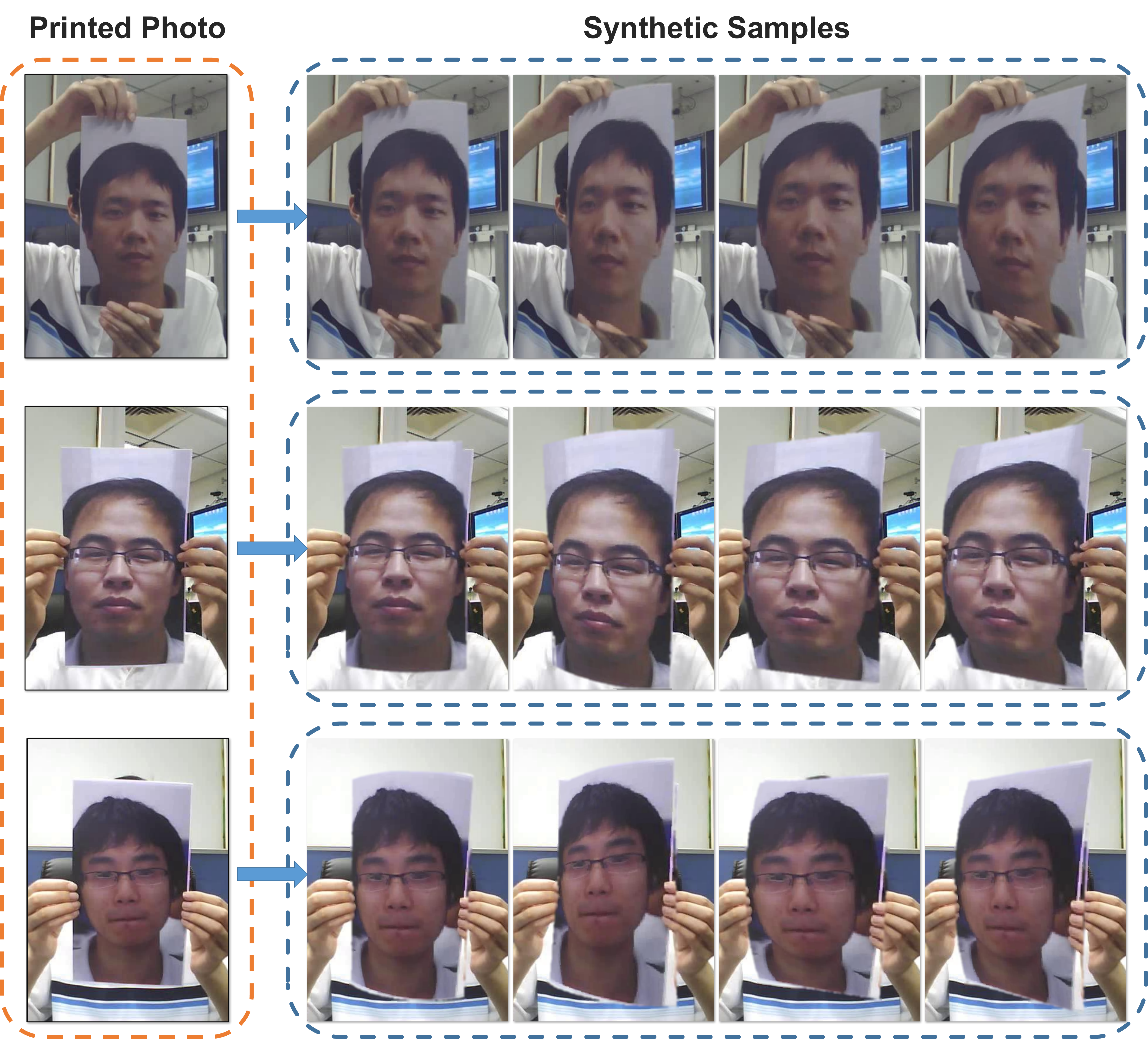}
  \caption{Virtual synthetic samples from CASIA-MFSD. Left column: the original printed photos. Right column: synthetic samples with different vertical/horizontal bending and rotating angles. These synthetic samples can significantly improve the face anti-spoofing performance.}
  \label{fig_demo}
\end{figure}

Print and replay attacks are the most common presentation attack (PA) ways and have been well studied in the academic field. Prior works can be roughly divided into three categories: cue-based, texture-based and deep learning based methods. Cue-based methods attempt to detect liveness motions~\cite{pan2007eyeblink,patel2016cross} such as eye blinking, lip, and head movements.
Texture-based methods aim to exploit discriminative patterns between live and spoof faces, by adopting hand-crafted features such as HOG and LBP. Deep learning based methods mainly consist of two types. The first one treats PA as a binary classification or pseudo-depth regression problem~\cite{li2016original,yang2013face,atoum2017face}. The other one tries to utilize temporal information of the video, such as applying the RNN-based structure~\cite{xu2015learning,liu2018learning}.


In practical applications, the replay attack can be easily detected using specialized sensors like depth or Near InfraRed (NIR) cameras, because the captured depth values of the spoof face lie in one flat surface, which is easily distinguished from live faces. While for the print attack, an imposter will try his best to fool the system, such as bending and rotating the printed photo.
However, most of the published databases miss the transformations, so that the trained models are easily spoofed by photo bending and rotating.

On the other hand, learning based methods especially deep learning based methods for face anti-spoofing need a large number of training samples to reduce overfitting. However, it is very expensive to acquire spoof data since the live faces should be re-printed and re-captured in many views.
 It is worth noting that, several recent works~\cite{guo2018face,su2015render,massa2016deep} have shown that synthesized images are effective for training CNN-based models in various tasks.

Motivated by these previous works, we propose to address these issues in a virtual synthesis manner. First, the high-fidelity virtual spoof samples with bending and out-of-plane rotating are synthesized through rendering from the transformed mesh in 3D space. Several synthetic examples are shown in Fig.~\ref{fig_demo}.
Second, deep models are trained on these synthetic samples with a data balancing method. To validate the effectiveness of our method, we design our experiments in two aspects: intra-database and inter-database testing. The intra-database testing is evaluated on CASIA-MFSD~\cite{zhang2012face} and Replay-Attack~\cite{chingovska2012effectiveness} for fair comparisons with other methods. For inter-database testing, we choose CASIA-MFSD~\cite{zhang2012face} as our training dataset and CASIA-RFS~\cite{lei2017countermeasures} as our testing set, since CASIA-RFS contain rotated and bent spoof faces series, which is more challenging. Besides, we rebuild the protocol on CASIA-RFS for better quantitative comparisons.

The main contributions of our work include:
\begin{itemize}
  \item A virtual synthesis method to generate bent and out-of-plane rotated spoofs is proposed. Large scales of spoof training data can be generated for training deep neural networks. 
  \item To train CNN from the large-scale synthetic spoof samples, a data balancing method is proposed to improve generalization of the face anti-spoofing model.
  \item We achieve the state-of-the-art performances on the CASIA-MFSD and Replay-Attack databases and obtain great improvement of generalization on the CASIA-RFS database.
\end{itemize}


\section{Related Work}
\label{sec_related_work}
We review related works from two perspectives: CNN-based methods for face anti-spoofing and data synthesis for CNN training.

\textbf{CNN-based Methods for Face Anti-spoofing.} Many CNN-based methods for face anti-spoofing have been recently proposed, which can be categorized into two groups: texture-based methods~\cite{li2016original,patel2016cross,li2016original,yang2014learn,xiao20193dma} and series-based methods~\cite{xu2015learning,liu2018learning}.

Most of the texture-based methods treat face anti-spoofing as a binary classification problem.
In ~\cite{li2016original,patel2016cross}, it uses the CaffeNet or VGG model pre-trained on ImageNet as initialization and then fine-tunes it on face-spoofing data. The SVM is finally applied for face spoofing detection. 
In ~\cite{li2016original}, different kinds of face features (e.g., multi-scale faces or hand-crafted features) are designed to feed into CNN.
In ~\cite{de2018learning}, a two-step training method is proposed to learn local and global features.
Recently, several studies~\cite{atoum2017face,liu2018learning,wang2018exploiting} indicate that the depth supervised based methods perform better than binary supervised. Atoum et al.~\cite{atoum2017face} propose to use the pseudo-depth map as supervised signals. A novel two-stream CNN-based approach for face anti-spoofing by extracting the local features and holistic depth maps from the face images is proposed. The fusion of the scores of two-steam CNNs leads to the final predicted class of live vs. spoof. Liu et al.~\cite{liu2018learning} fuse the estimated depth and the rPPG signals to distinguish live vs. spoof faces. They argue that auxiliary supervision such as pseudo-depth map and rPPG signals are important to guide the learning toward discriminative and generalizable cues. One of the most recent work~\cite{wang2018exploiting} proposes a depth supervised face anti-spoofing model in both spatial and temporal domains thus more robust and discriminative features can be extracted to classify live and spoof faces.

Series-based methods aim to fully utilize the temporal information of serialized frames. Feng et al.~\cite{feng2016integration} feed both optical flow map and Shearlet feature to CNN. The work~\cite{xu2015learning} proposes a deep neural network architecture combining Long Short-Term Memory (LSTM) units with CNN. In~\cite{gan20173d}, 3D convolution network is adopted in short video frame level to distinguish live vs. spoof face. Besides, the rPPG signals extracted from serialized frames are used as auxiliary supervision for classification in~\cite{liu2018learning}.

\textbf{Virtual Synthesis for CNN Training.} Generally, CNN-based methods need a large scale of training data to reduce overfitting. However, training data is difficult to collect in many cases. Several recent works focus on creating synthetic images to augment the training data, \eg face alignment~\cite{zhu2017face,guo2020towards}, face recognition~\cite{guo2018face}, 3D human pose estimation~\cite{chen2016synthesizing,du2016marker,ghezelghieh2016learning,rogez2016mocap}, pedestrian detection~\cite{pishchulin2012articulated,pishchulin2011learning} and action recognition~\cite{rahmani2015learning,rahmani20163d}. Zhu et al.~\cite{zhu2016face,zhu2017face} attempt to utilize 3D information to synthesize face images in large poses to provide abundant samples for training. Similarly, Guo et al.~\cite{guo2018face} synthesize high-fidelity face images with eyeglasses as training data based on 3D face model and 3D eyeglasses and achieve better face recognition performance on real eyeglass face testset. Pishchulin et al.~\cite{pishchulin2011learning} generate synthetic images with a game engine. In~\cite{pishchulin2012articulated}, they deform 2D images with a 3D model. In~\cite{rahmani2015learning}, action recognition is addressed with synthetic human trajectories from MoCap data.

\section{3D Virtual Synthesis}
\label{sec_3d_virtual_synthesis}
In this section, we demonstrate how to synthesize virtual spoof face data.
The purpose of synthesis is trying to simulate the behaviors of bending and out-of-plane rotating.

\subsection{3D Meshing and Deformation}
\begin{figure}[htb]
  \centering 
  
  \subfigure[]{
  \label{fig_image_meshing_a}
  \includegraphics[height=0.2\textwidth]{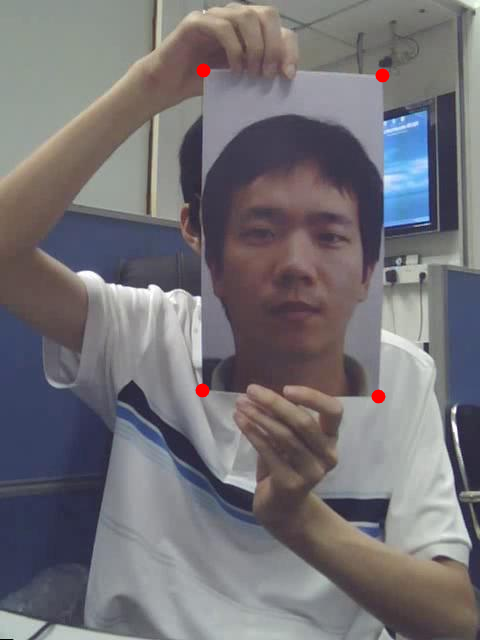}}
  \subfigure[]{
  \label{fig_image_meshing_b}
  \includegraphics[height=0.2\textwidth]{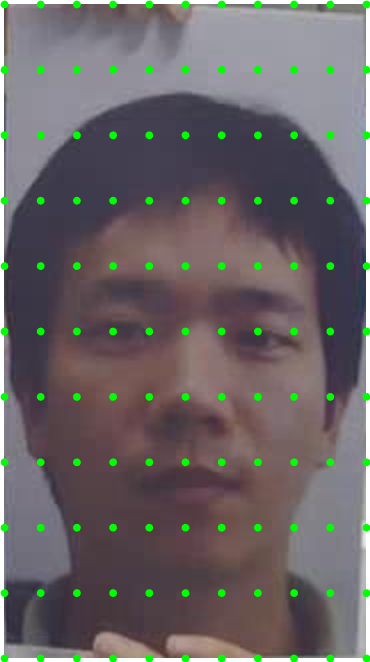}}
  \subfigure[]{
  \label{fig_image_meshing_c}
  \includegraphics[height=0.2\textwidth]{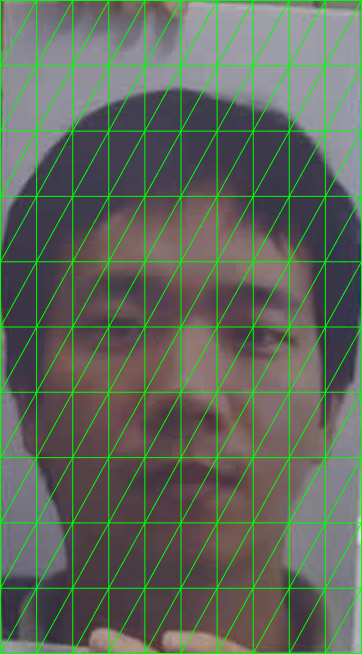}}
  \subfigure[]{
  \label{fig_image_meshing_d}
  \includegraphics[width=0.22\textwidth]{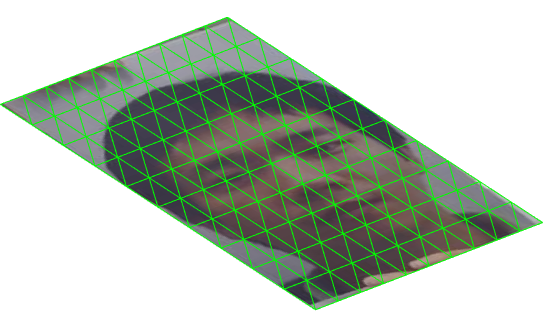}}
  \subfigure[]{
  \label{fig_image_meshing_e}
  \includegraphics[width=0.22\textwidth]{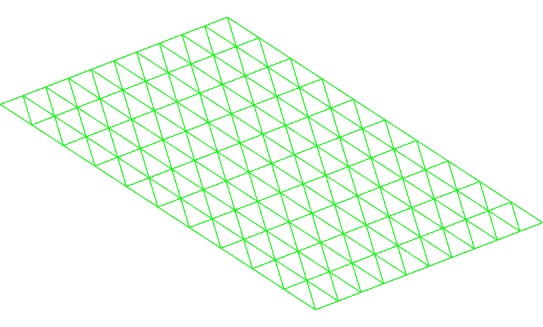}}
  
  \caption{3D Image Meshing. (a) The input source image marked with corner anchors. (b) The cropped image and sampled anchors. (c) The triangle mesh overlapped with the cropped image (2D view). (d) The triangle mesh overlapped with cropped image (3D view) (e) 3D view of the triangle mesh.}
  \label{fig_image_meshing}
\end{figure}

We assume that a printed photo has a simple 3D structure, which is a flat surface so that we can easily transfer the printed photo to a 3D object and manipulate its appearance in 3D space.
First, we label four corner anchors (Fig.~\ref{fig_image_meshing_a}) to crop the printed photo region. Second, the anchors are uniformly sampled on cropped region (Fig.~\ref{fig_image_meshing_b}). Their depths are set the same since we treat the printed photo as a plane. Finally, the delaunay algorithm is applied to triangulate these anchors to mesh the printed photo into a virtual 3D object (Fig.~\ref{fig_image_meshing_c}). The 3D view of virtual 3D printed photo is are shown in Fig.~\ref{fig_image_meshing_d} and Fig.~\ref{fig_image_meshing_e}.

After 3D meshing, the 3D transformation operations such as rotating and bending can be applied. We use a rotation matrix $R$ to rotate the 3D object. Let $V_{xyz}$ representing the sampled anchors, the rotating operation can be represented by $V = R * V_{xyz}$.

\begin{figure}[!htb]
  \small
  \centering
  \includegraphics[width=0.275\textwidth]{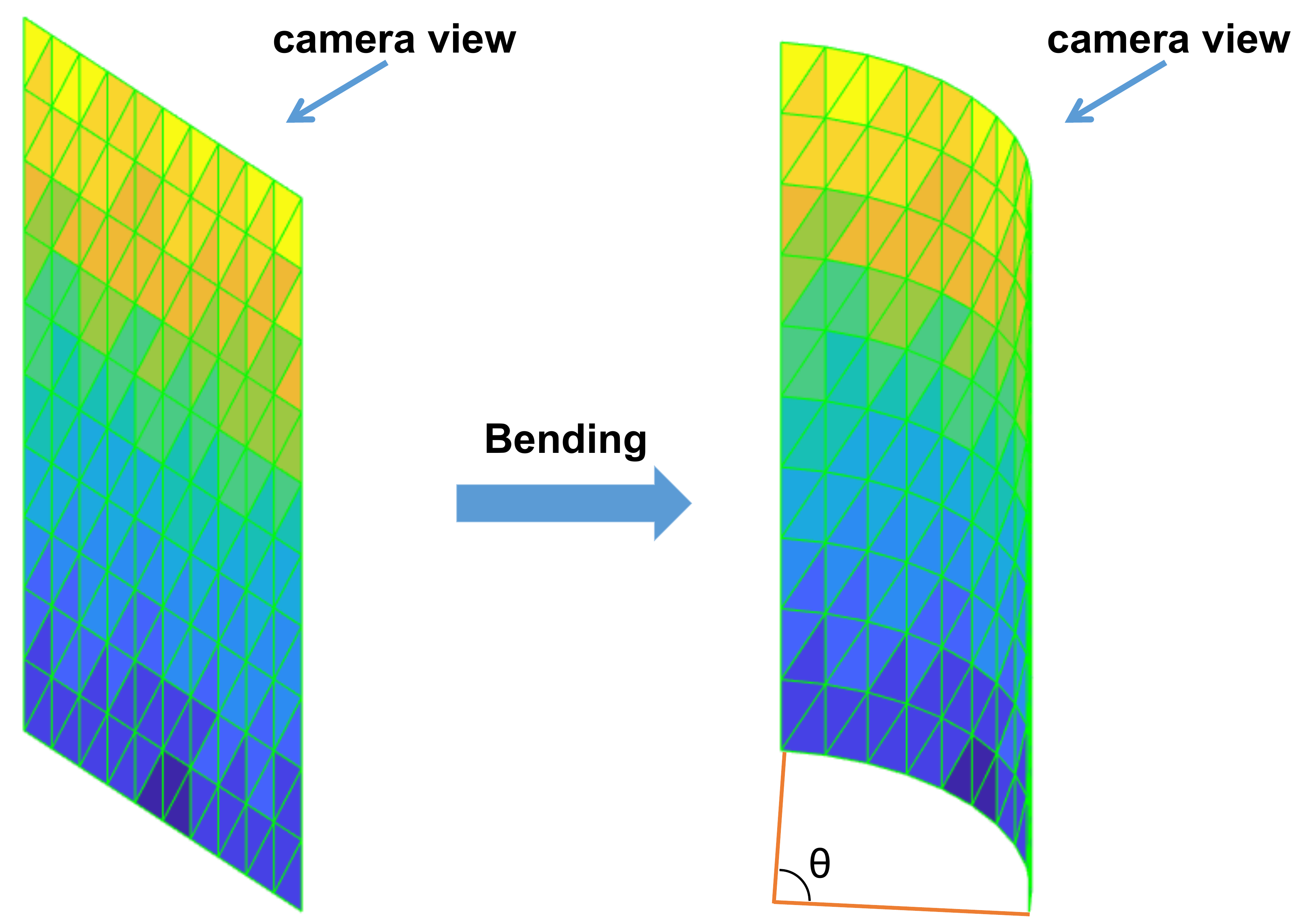}
  \caption{Vertical bending of the planar triangle mesh. Left is 3D planar mesh and right is 3D bending mesh.}
  \label{fig_bending}
\end{figure}

\begin{figure}[!htb]
  \centering
  \includegraphics[width=0.3\textwidth]{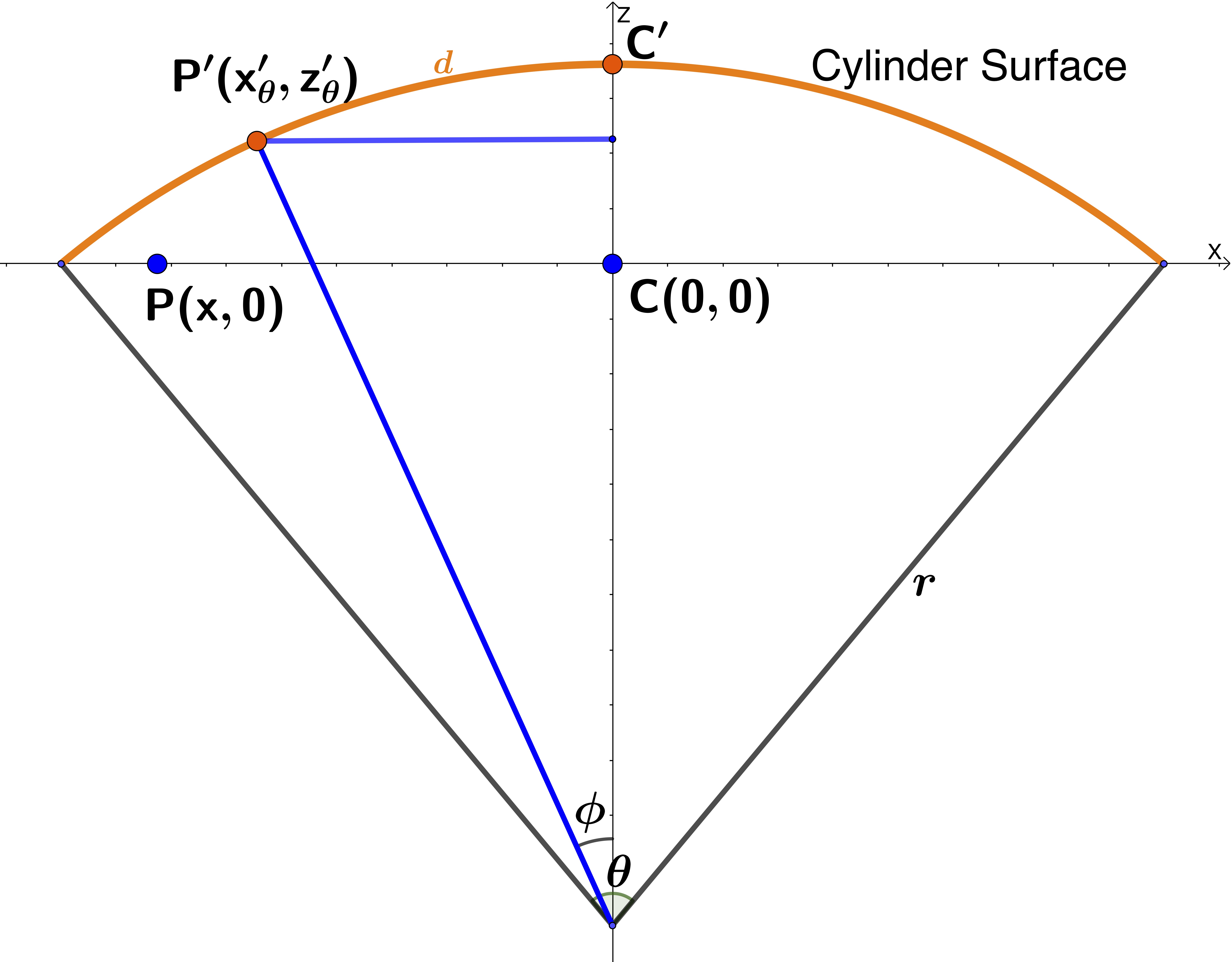}
  \caption{2D aerial view of vertical bending.}
  \label{fig_cylinder}
\end{figure}

The bending operation involves non-rigid transformation. In order to simulate the bending operation, we deform the 3D planar mesh to a cylinder, with the length along the horizontal and vertical directions preserved. 
As shown in Fig.~\ref{fig_bending}, the bending angle $\theta$ measures the degree of vertical bending.
We show the 2D aerial view of vertical bending in Fig.~\ref{fig_cylinder}, where $P$ is an anchor point on original 3D planar mesh and $P'$ is the deformed anchor point on cylinder surface. The radius of the circle on cylinder mesh surface can be first calculated by $r = l / \theta$, where $l$ denotes the width of 3D planar mesh. The radian of arc $P'C'$ can be calculated by $\phi = d / l \cdot \theta$, where $d = x - l / 2$ is the distance from anchor point $P$ to the mesh center $C$ (or from $P'$ to $C'$). The new position of anchor $P'$ can be formulated as follows:
\begin{equation}
	\begin{aligned}
		x'_{\theta} &= r \sin{\phi} = \frac{l}{\theta} \cdot \sin{(\frac {x - l / 2} {l} \cdot \theta)}, \\
		z'_{\theta} &= r \cos{\phi} - r \cos{\frac {\theta} 2} \\ 
		&= \frac{l}{\theta} \cdot \big [ \cos{(\frac {x - l / 2} {l} \cdot \theta)} - \cos{\frac \theta  2} \big],
	\end{aligned}
\end{equation}
where $\theta$ is the bending angle. Horizontal bending is similar to vertical bending. Besides, the rotating and bending can be composed to synthesize more varied samples.

\subsection{Perspective Projection}
When a printed photo is captured, the object in the distance should appear smaller than the object close by.
However, this effect is always ignored and the weak perspective projection is adopted in many virtual synthesis techniques~\cite{guo2018face,zhu2017face}.
In this work, we use perspective projection for more realistic synthesis.

To perform the perspective transformation, we must first approximate the physical size of the printed photo.
We assume the pixel distance and real distance between two eyes centers as $d_p$ and $d_r$ respectively, then the scale factor of the transformation from image space to world coordinated system can be approximated as:
\begin{equation}
	s = d_p / d_r,
\end{equation}
where $s$ is the scale factor of transformation from image space to world coordinate system. Then the perspective projection can be applied by
\begin{equation}
	\begin{aligned}
		b_x &= v_x \cdot (f / d_z), \\
		b_y &= v_y \cdot (f / d_z).
	\end{aligned}
\end{equation}
where $f$ is the focal length, $d_z$ indicates the physical depth distance from camera to photo, $v_x$, $v_y$ are anchor vertices in world coordinate system, and $b_x$, $b_y$ are projected 2D vertices by perspective transformation.
For convenience, the real distance between eyes centers $d_r$, the focal length $f$ and the depth $d_z$ are approximated by constant values based on prior.

\begin{figure}[htb]
  \centering 
  \subfigure[]{
  \label{fig_image_ori}
  \includegraphics[height=0.2\textwidth]{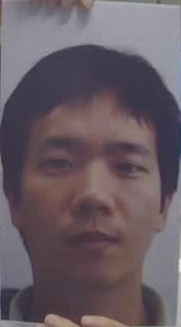}}
  \subfigure[]{
  \label{fig_image_weak_projection}
  \includegraphics[height=0.2\textwidth]{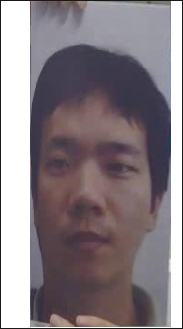}}
  \subfigure[]{
  \label{fig_image_perspective_projection}
  \includegraphics[height=0.2\textwidth]{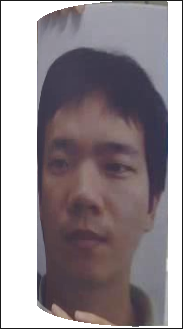}}
  
  \caption{Different projections. (a) Texture of the cropped printed photo region. (b) Rendered texture by weak perspective projection. (c) Rendered texture by perspective projection.}
  \label{fig_projection}
\end{figure}

After the mesh deformation and projection, synthetic samples can be rendered by Z-buffer. Fig.~\ref{fig_projection} shows the difference between weak perspective projection and perspective projection. The image synthesized by perspective projection in Fig.~\ref{fig_image_perspective_projection} is more realistic than the one by weak perspective projection in Fig.~\ref{fig_image_weak_projection}.

\subsection{Post-processing}
\begin{figure}[htb]
  \centering
  \subfigure[]{
  \label{fig_post_process_before}
  \includegraphics[width=0.16\textwidth]{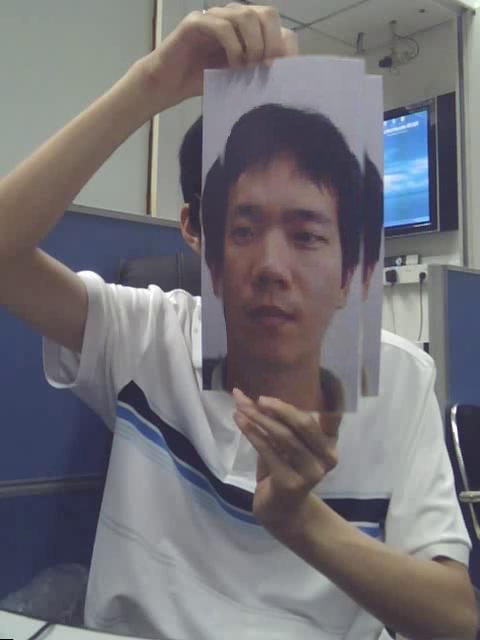}}
  \subfigure[]{
  \label{fig_post_process_corner_adjust}
  \includegraphics[width=0.16\textwidth]{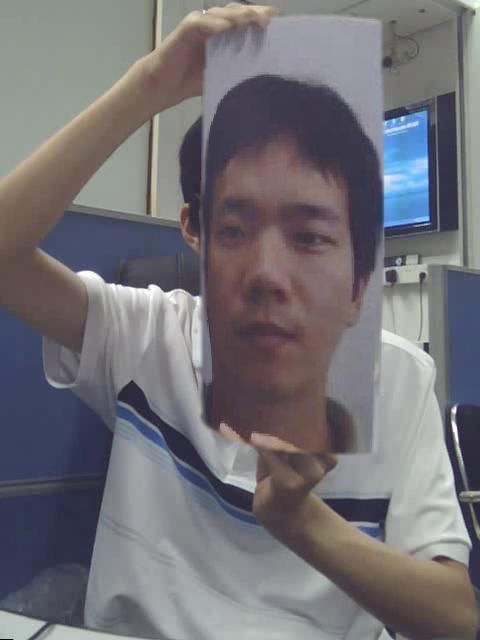}}
  
  \caption{Post-processing of the synthetic printed photo. (a) Fused image without full overlapping. (b) Fused image with full overlapping.}
  \label{fig_post_process}
\end{figure}

Due to the mesh deformation and perspective projection, the size of synthetic photos will be changed (see Fig.~\ref{fig_post_process_before}). To improve the fidelity of the final synthetic sample, we try to make sure the synthetic printed photo fully overlaps the originally printed photo region (see Fig.~\ref{fig_post_process}).
Besides, the gaussian image filter is applied to make the fused edge smoother. Several final synthetic spoof results are shown in Fig.~\ref{fig_demo}.

\section{Deep Network Training on Synthetic Data}
\label{sec_training}
\begin{figure}[!htb]
  \centering
  \small
  \includegraphics[width=0.45\textwidth]{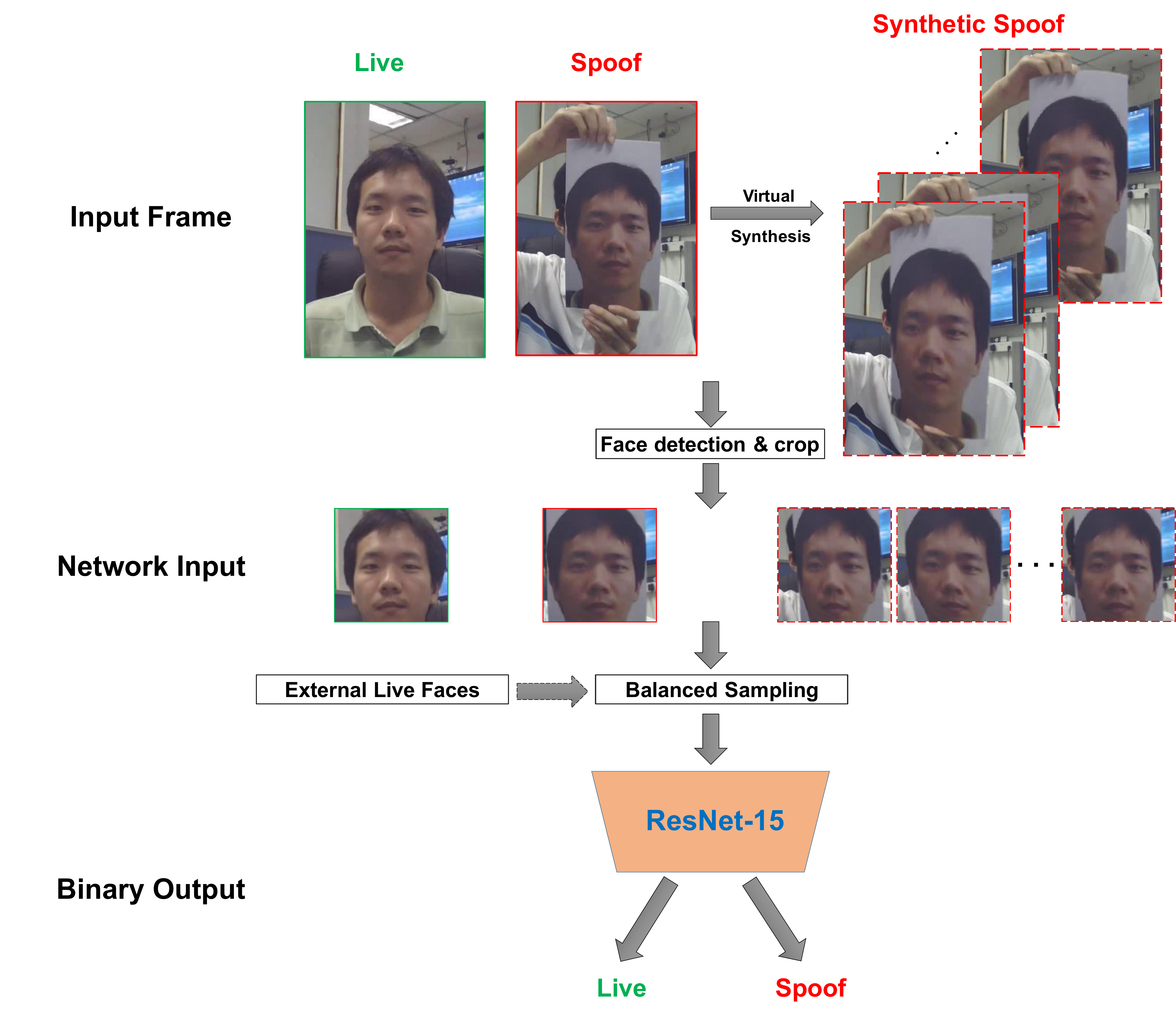}
  \caption{The pipeline of our proposed deep network training with synthetic data.}
  \label{fig_pipeline}
\end{figure}

In this section, we present our approach to training from synthetic spoof data. Fig.~\ref{fig_pipeline} shows a high-level illustration of our training pipeline. Virtual synthetic spoof samples are used to supervise the network training with our data balancing methods.

\textbf{Data Balancing.} We can generate as many spoof samples as we want by the synthesis method, but the amount of live samples is fixed. As a result, the CNN model trained has a bias to virtual spoofs due to the imbalance of live and virtual spoof samples.

There are two methods to mitigate the impact of data imbalance: balanced sampling strategy and importing external live samples.
To perform balanced sampling, the ratio of sampled live and spoof instances in each min-batch is kept fixed during training after augmenting dataset.
Besides, the live samples are much easier to acquire than the spoof ones in practical applications, such as the samples from face recognition databases~\cite{gross2010multi,kemelmacher2016megaface}. Therefore, unlimited external live samples can be imported to balance the distribution of training data. 

\textbf{How to Treat Synthetic Samples.}
The previous synthesis-based work~\cite{chen2016synthesizing} first uses the synthetic samples to train one initialization model and then fine-tunes it on the real data.
Since our synthesis is high-fidelity and possesses much more variations than the real data, we treat equally the synthetic spoof samples with the real ones and directly train models on the joint data of synthetic and real data.

\section{Experiments}
\label{exp}
We evaluate the effectiveness of learning with virtual synthetic data from two aspects: CASIA-MFSD~\cite{zhang2012face} and Replay-Attack~\cite{chingovska2012effectiveness} for intra-database evaluation and CASIA-RFS~\cite{lei2017countermeasures} for inter-database evaluation.
Performance evaluations on CASIA-MFSD and Replay-Attack databases are for fair comparisons with other previous methods.
The CASIA-RFS database is much more challenging since it contains rotated live and spoof faces with various poses. We use it as the testing set and the entire CASIA-MFSD database as the training set to carry out the inter-database testing.

\subsection{Database and Protocols}
\textbf{CASIA-MFSD.} This database contains 50 subjects, and 12 videos for each subject under different resolutions and light conditions. Three different spoof attacks are designed: replay, warp print and cut print attacks. The database contains 600 video recordings, in which 240 videos of 20 subjects are used for training and 360 videos of 30 subjects for testing.

\textbf{Replay-Attack.} This database consists of 1,300 videos from 50 subjects. These videos are collected under controlled and adverse conditions and are divided into training, development and testing sets with 15, 15 and 20 subjects respectively.

\textbf{CASIA-RFS.} This database contains 785 videos for genuine faces and 1950 videos for spoof faces. The videos are collected using three devices with different resolutions: digital camera (DC), mobile phone (MP) and web camera (WC). Two kinds of spoof attacks including planar and bent photo attacks are designed.

\textbf{Protocols.} For CASIA-MFSD and Replay-Attack databases, our experiments follow the associated protocols, the EER and HTER indicators are reported. The Replay-Attack database has already provided a development set thus our HTER threshold is determined on it. As for inter-database evaluation, we build the protocols following~\cite{zinelabidine2017oulunpu}. Attack Presentation Classification Error Rate (APCER), Bona Fide Presentation Classification Error Rate (BPCER), Average Classification Error Rate (ACER), Top-1 accuracy are all evaluated. Particularly, the APCER here represents the highest error among plane printed and bending printed attacks.

\subsection{Implementation Details}

\textbf{Network Structure.} Our network structure is modified from ResNet~\cite{he2016deep}. The input size of original ResNet is 224 $\times$ 224, while ours is 120 $\times$ 120. As a result, the original 7 $\times$ 7 convolution in the first layer is replaced by 5 $\times$ 5 and follows one 3 $\times$ 3 convolution layer to preserve the dimension of feature map output. Finally, one 15 layers ResNet (ResNet-15) structure is designed for our task and shown in Table~\ref{tab_resnet15}.

\begin{table}[t]
\small
\centering
\caption{Our ResNet-15 network structure. Conv3.x, Conv4.x, and Conv5.x indicate convolution units which contain multiple convolution layers, and residual blocks are shown in double-column brackets. For instance, [3 $\times$ 3, 128] $\times$ 3 denotes 3 cascaded convolution layers with 128 feature maps with filters of size 3 $\times$ 3, and S2 denotes stride 2.}
\label{tab_resnet15}
\begin{tabular}{c|c}
\hline
\textbf{Layers} & \textbf{15-layer CNN} \\
\hline
Conv1.x & [5$\times$5, 32]$\times$1, S2 \\ \hline
Conv2.x & [3$\times$3, 64]$\times$1, S1 \\ \hline
Conv3.x & $\left[\begin{aligned}&3\times3, 128\\&3\times3, 128\end{aligned}\right]\times 2$, S2 \\ \hline
Conv4.x & $\left[\begin{aligned}&3\times3, 256\\&3\times3, 226\end{aligned}\right]\times 2$, S2 \\ \hline
Conv5.x & $\left[\begin{aligned}&3\times3, 512\\&3\times3, 512\end{aligned}\right]\times 2$, S2 \\ \hline
Global Pooling \& FC & 512 \\ \hline
\end{tabular}
\end{table}

\textbf{Training.} Our experiments are based on PyTorch framework and GeForce GTX TITAN X GPU devices. All training images are cropped and aligned to the size of 120 $\times$ 120 by similar transformation, then being normalized by subtracting 127.5 and being divided by 128. For all three databases, we use SGD with a mini-batch size of 64 to optimize the network, with the weight decay of 0.0005 and momentum of 0.9. Image horizontal flipping is adopted as standard augmentation. The weight parameters of ResNet-15 model are randomly initialized. We train 30 epochs for each experiment. We set the initial learning rate of 0.1, then decrease it by multiplying 0.1 in 10th and 20th epoch respectively.

The ratios of the number of live/spoof samples on the CASIA-MFSD and Replay-Attack databases are all about 1:3. We keep this ratio in each mini-batch sampling on augmented training data when applying balanced sampling. More comparisons can be referred to Sec.\ref{exp_comparison}. For inter-database testing in Sec.\ref{sec_inter_database}, we use part of CMU Multi-PIE Face Database (MultiPIE)~\cite{gross2010multi}. Totally, 8,120 face images with various poses are introduced as external data.

\begin{figure}[!htb]
  \centering
  \includegraphics[width=0.4125\textwidth]{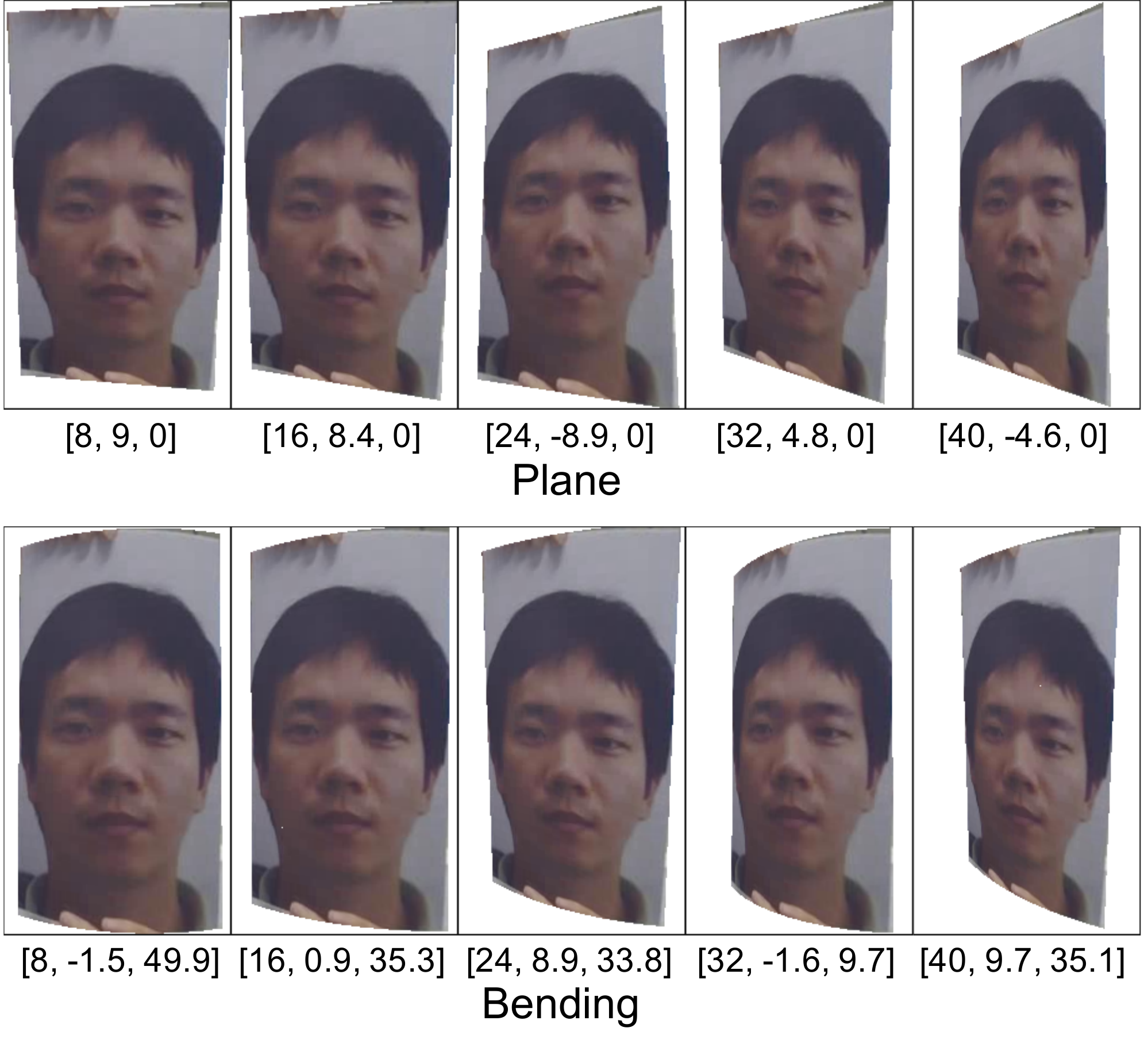}
  \caption{Synthetic samples before post-processing. The first row: five synthesized planar samples. The second row: five synthesized vertically bent samples. The values in brackets indicate yaw, pitch, and bending angle successively.}
  \label{fig_batch}
\end{figure}

\textbf{Synthesis.} For each printed spoof instance in CASIA-MFSD and Replay-Attack, we generate ten synthetic samples, in which five samples are rotated and bent, another five ones are only rotated. For each replay spoof sample, we generate five synthetic samples without bending, since the replay devices cannot be bent.
During rotating, the yaw angle is uniformly drawn from the interval [0, 40], the pitch angle is from [-10, 10] and the bending angle is from [30, 60]. We show one example in Fig.~\ref{fig_batch}.

\begin{table}[t!]
	\caption{ACER(\%), Top-1(\%), EER(\%) and HTER(\%) on CASIA-MFSD with different projections.}	
	\label{tab_projections}
\small
\centering
	\begin{tabular}{c|c|c|c|c}
		\hline
		\multirow{2}{*} {\textbf{Projection}} & \textbf{ACER} & \textbf{Top-1} & \textbf{EER} & \textbf{HTER} \\
		 & (\%) & (\%) & (\%) & (\%) \\ \hline 
		Weak Perspective & 3.33 & 97.78 & 2.59 & 2.41 \\ 
		Perspective & 2.22 & 98.61 & 2.22 & 1.67 \\ \hline
	\end{tabular}
\end{table}



\begin{table}[]
\centering
\caption{ACER(\%), Top-1(\%), EER(\%) and HTER(\%) on CASIA-MFSD and Replay-Attack. Syn indicates synthetic data and BS indicates the balanced sampling.}
\label{tab_replay_attack_casia_ablation}
\resizebox{0.45\textwidth}{!} {
\begin{tabular}{c|c|c|c|c|c}
\hline
\textbf{Database} & \textbf{Method} & \begin{tabular}[c]{@{}c@{}}\textbf{ACER}\\ (\%)\end{tabular} & \begin{tabular}[c]{@{}c@{}}\textbf{Top-1}\\ (\%)\end{tabular} & \begin{tabular}[c]{@{}c@{}}\textbf{EER}\\ (\%)\end{tabular} & \begin{tabular}[c]{@{}c@{}}\textbf{HTER}\\ (\%)\end{tabular} \\ \hline
\multirow{3}{*}{CASIA-MFSD} & Baseline & 5 & 96.67 & 4.44 & 3.89 \\ 
 & Syn w/o BS & 4.44 & 97.78 & 3.33 & 2.78 \\ 
 & Syn w/ BS & 2.22 & 98.61 & 2.22 & 1.67 \\ \hline
\multirow{3}{*}{Replay-Attack} & Baseline & 4.17 & 98.13 & 2.50 & 3.50 \\ 
 & Syn w/o BS & 2.50 & 98.14 & 1.25 & 1.75 \\ 
 & Syn w/ BS & 0.21 & 99.79 & 0.25 & 0.63 \\ \hline
\end{tabular}
}
\end{table}


\begin{table}[t!]
	\caption{APCER(\%), BPCER(\%), ACER (\%) and Top-1(\%) on CASIA-RFS. Syn indicates synthetic data.}
	\label{tab_casia_rfs_ablation}
	\small
	\centering
\resizebox{0.45\textwidth}{!} {
	\begin{tabular}{c|c|c|c|c}
	\hline
	\multirow{2}{*} {\textbf{Method}} & \textbf{APCER} & \textbf{BPCER} & \textbf{ACER} & \textbf{Top-1} \\
		 & (\%) & (\%) & (\%) & (\%) \\ \hline 
    Baseline & 20.96 & 23.98 & 22.47 & 81.83 \\ 
	Syn w/o MultiPIE & 4.01 & 50.23 & 27.12 & 82.63 \\ 
	Syn w/ MultiPIE & 4.68 & 18.75 & 11.72 & 91.68 \\ \hline
	\end{tabular}
}
\end{table}


\subsection{Experimental Comparison}
\label{exp_comparison}
\subsubsection{Ablation Study} 
\label{sec_abalation_study}
We evaluate the performance of different projections, synthetic data, and data balancing in this section.

\textbf{Projection.} We explore the weak perspective and perspective projections of the spoof data synthesis. From results in Table~\ref{tab_projections}, we can see that the perspective projection has a significant improvement over the weak perspective projection with an EER $2.22\%$ and an HTER $1.67\%$. Moreover, ACER improves from $3.33\%$ to $2.22\%$ and Top-1 accuracy rises from $97.78\%$ to $98.61\%$. The results indicate that perspective projection is more suitable for virtual spoof data synthesis.

\textbf{Synthetic Data \& Balanced Sampling (BS).} For CASIA-MFSD and Replay-Attack in Table~\ref{tab_replay_attack_casia_ablation}, we compare three configurations: (i) the original dataset is used without the synthetic spoof data and BS; (ii) the synthetic spoof data is added; (iii) both the synthetic data and BS are adopted. The results in Table~\ref{tab_replay_attack_casia_ablation} indicate that (ii) is considerably better than (i) in ACER, Top-1 accuracy, EER, and HTER on both CASIA-MFSD and Replay-Attack. It shows the synthetic spoof data is effective for training CNN. Besides, (iii) achieves the best performance on both CASIA-MFSD and Replay-Attack, which validates the effectiveness of our BS and the advantage of combining synthetic samples and BS.


\textbf{External Live Samples.} For CASIA-RFS in Table~\ref{tab_casia_rfs_ablation}, we compare three configurations: (i) neither synthetic data nor MultiPIE is used (baseline); (ii) the synthetic spoof data is used; (iii) both MultiPIE and the synthetic spoof data are used. The balanced sampling is applied in (ii) and (iii).
Compared with the baseline, the addition of synthetic data makes APCER decrease from 20.96\% to 4.01\%, but BPCER becomes higher. It shows that the spoof face classification accuracy gets better but the live face classification precision drops. We think it is because live faces in CASIA-RFS have a wide range of poses, while such variation in CASIA-MFSD is limited.
Once MultiPIE is used for training, BPCER drops obviously from 50.23\% to 18.75\% and APCER almost unchanges.
Finally, the Top-1 and EER indicators on CASIA-RFS get improved by 9.85\% and 9.67\% respectively compared with the baseline.
The above results show that external live data, \eg MultiPIE, improves the generalization by a large margin.


\begin{table}[t!]
	\small
	\centering
	\caption{EER (\%) and HTER (\%) on CASIA-MFSD. BS indicates the balanced sampling strategy.}
	\label{tab_casia_mfsd}
	\begin{tabular}{c|c|c}
	\hline
	\textbf{Method} & \textbf{EER (\%)} & \textbf{HTER (\%)} \\ \hline
	Fine-tuned VGG-Face~\cite{li2016original} & 5.20 & - \\ 
	DPCNN~\cite{li2016original} & 4.50 & - \\ 
	Multi-Scale~\cite{yang2014learn} & 4.92 & - \\ 
	CNN~\cite{xu2015learning} & 6.20 & 7.34 \\ 
	LSTM-CNN~\cite{xu2015learning} & 5.17 & 5.93 \\ 
	YCbCr+HSV-LBP~\cite{boulkenafet2015face} & 6.20 & - \\
	Feature Fusion~\cite{siddiqui2016face} & 3.14 & -\\ 
	Fisher Vector~\cite{boulkenafet2017face} & 2.80 & -\\ 
	Patch-based CNN~\cite{atoum2017face} & 4.44 & 3.78 \\ 
	Depth-based CNN~\cite{atoum2017face} & 3.78 & 2.52 \\ 
	Patch\&Depth Fusion~\cite{atoum2017face} & 2.67 & 2.27 \\ \hline
	
	\textbf{Ours (Syn w/ BS)} & \textbf{2.22} & \textbf{1.67} \\ \hline
	\end{tabular}
\end{table}

\begin{table}[t!]
	\small
	\centering
	\caption{EER (\%) and HTER (\%) on Replay-Attack. BS indicates the balanced sampling strategy.}
	\label{tab_replay_attack}
	\begin{tabular}{c|c|c}
	\hline
	\textbf{Method} & \textbf{EER (\%)} & \textbf{HTER (\%)} \\ \hline
	Fine-tuned VGG-Face~\cite{li2016original} & 8.40 & 4.30 \\ 
	DPCNN~\cite{li2016original} & 2.90 & 6.10 \\ 
	Multi-Scale~\cite{yang2014learn} & 2.14 & - \\ 
	YCbCr+HSV-LBP~\cite{boulkenafet2015face} & 0.40 & 2.90 \\ 
	Fisher Vector~\cite{boulkenafet2017face} & 0.10 & 2.20 \\ 
	Moire pattern~\cite{patel2015live} & - & 3.30 \\ 
	Patch-based CNN~\cite{atoum2017face} & 4.44 & 3.78 \\ 
	Depth-based CNN~\cite{atoum2017face} & 3.78 & 2.52 \\ 
	Patch\&Depth Fusion~\cite{atoum2017face} & 0.79 & 0.72 \\ 
	FASNet~\cite{lucena2017transfer} & - & 1.20 \\ \hline 
	\textbf{Ours (Syn w/ BS)} & \textbf{0.25} & \textbf{0.63} \\ \hline
	\end{tabular}
\end{table}

\subsubsection{Intra-database Testing}
The intra-database testing is performed on CASIA-MFSD and Replay-Attack.
Table~\ref{tab_casia_mfsd} shows the comparisons of our proposed synthesis-based method with state-of-the-art methods.
As shown in Table~\ref{tab_casia_mfsd}, our synthesis-based method outperforms other methods in both EER and HTER.
For Replay-Attack database, we perform comparisons in Table~\ref{tab_replay_attack}. We can see the proposed method also outperforms other state-of-the-art methods. Though our method has similar EER with several methods, the HTER is smaller than theirs, which means we have lower false acceptance and false rejection rates.

\begin{table}[!t]
\caption{APCER (\%), BPCER (\%), ACER(\%) and Top-1(\%) on CASIA-RFS for inter-database testing.}
\label{tab_casia_rfs}
\small
\centering
\resizebox{0.45\textwidth}{!} {
\begin{tabular}{c|c|c|c|c}
\hline
	\multirow{2}{*} {\textbf{Method}} & \textbf{APCER} & \textbf{BPCER} & \textbf{ACER} & \textbf{Top-1} \\
		 & (\%) & (\%) & (\%) & (\%) \\ \hline 
	FASNet~\cite{lucena2017transfer} & 1.23 & 47.27 & 24.25 & 84.48 \\ 
	Patch-based CNN~\cite{atoum2017face} & 12.93 & 29.89 & 21.41 & 83.54 \\ 
	Baseline & 20.96 & 23.98 & 22.47 & 81.83 \\ \hline 
	\textbf{Ours} & \textbf{4.68} & \textbf{18.75} & \textbf{11.72} & \textbf{91.68} \\ \hline
\end{tabular}
}
\end{table}

\subsubsection{Inter-database Testing}
\label{sec_inter_database}
We perform the inter-database testing on CASIA-RFS, in which CASIA-MFSD is used for training. In Table~\ref{tab_casia_rfs}, we compare our method with other CNN-based methods and our baseline (no additional synthetic data or external live data). The other CNN-based methods are re-implemented following the descriptions in original papers. The results show that our method outperforms other CNN-based methods and the baseline model. The inter-database testing also validates the effectiveness of synthetic spoof data and the data balancing strategy.

\section{Conclusion}
\label{sec_conclusion}
In this paper, we have shown successful large-scale training of CNNs from synthetically generated spoof data. For the data imbalance brought by the spoof data, we exploit two methods for balancing it: balanced sampling and adding external live samples. Experimental results show that our synthetic spoof data and data balancing methods greatly promote the performance for face anti-spoofing. The promising performance shows the great potential for advancing face anti-spoofing using large-scale synthetic data. Besides, more realistic and more kinds of spoof synthesis are future directions.

\section*{Acknowledgment}
This work was supported by the Chinese National Natural Science Foundation Projects \#61876178, \#61806196, \#61872367, \#61572501, \#61806203.


{\small
\bibliographystyle{ieee}
\bibliography{egbib}
}

\end{document}